# Genetic algorithms with DNN-based trainable crossover as an example of partial specialization of general search

Alexey Potapov[1,2,3], Sergey Rodionov[3,4]

[1]CT Lab, ITMO University, St. Petersburg, Russia
[2]St. Petersburg State University, St. Petersburg, Russia
[3]AIDEUS, Russia
[4]Aix Marseille Université, CNRS, LAM (Laboratoire d'Astrophysique de Marseille) UMR 7326, 13388, Marseille, France
{pas.aicv, astroseger}@gmail.com

**Abstract.** Universal induction relies on some general search procedure that is doomed to be inefficient. One possibility to achieve both generality and efficiency is to specialize this procedure w.r.t. any given narrow task. However, complete specialization that implies direct mapping from the task parameters to solutions (discriminative models) without search is not always possible. In this paper, partial specialization of general search is considered in the form of genetic algorithms (GAs) with a specialized crossover operator. We perform a feasibility study of this idea implementing such an operator in the form of a deep feedforward neural network. GAs with trainable crossover operators are compared with the result of complete specialization, which is also represented as a deep neural network. Experimental results show that specialized GAs can be more efficient than both general GAs and discriminative models.

## 1    Introduction

Solomonoff's theory of universal induction [1] has been ignored by the machine learning community for a long time because of its impracticality. However, one can find an apparent (although not explicitly declared) trend towards the universal induction in some recent works coming from the mainstream approaches in machine learning. For example, such deep learning models as Neural Turing Machine [2], Differentiable Neural Computer [3], Differentiable Forth Interpreter [4], Neural Programmer-Interpreter [5] and others are designed directly to perform inference in the space of algorithms that is the main feature of universal induction. The probabilistic programming field features the development of Turing-complete languages with general inference engines for arbitrary generative models. Using such inference engine, one can obtain a sort of universal induction algorithm by making inference on a model that generates arbitrary programs.

However, these efforts encounter some difficulties with scaling to the inference of non-trivial algorithms. These difficulties and the impracticality of the basic universal induction have the same origin. Indeed, the search in the Turing-complete space is very difficult, and general methods are not able to perform it efficiently or effectively. Deep neural networks heavily rely on the gradient descent, which application to the differential embedding of algorithms with sequential nature is prone to converge to inaccurate solutions [6]. The works on probabilistic programming languages (PPLs) are much more focused on evaluating posterior probabilities over all solutions, and frequently even don't consider the search problem utilizing simple enumeration or random search techniques.

Possibility to solve the universal induction problem with one simple and efficient method is doubtful. On the other hand, any fixed set of practical machine learning methods that work in Turing-incomplete model spaces is insufficient for the needs of artificial general intelligence. One general idea how to avoid these two undesirable extremes is meta-learning or, more generally, meta-inference, i.e. inference of new task-specific inference or learning algorithms. Meta-inference algorithms can both be computationally feasible and produce new efficient narrow inference algorithms.

Meta-learning including learning efficient forms of gradient descent [7] and more specific reinforcement learning algorithms [8, 9] in the deep learning framework has become quite popular recently reincarnating and developing further the ideas formulated earlier [10]. There are also probabilistic programming systems (e.g. [11]), which inference engines adapt to the given generative model (program), and are automatically reduced to the efficient inference methods, if the model falls into some narrow class (e.g. a form of message-passing algorithms on factor graphs). However, all these results are not put into the context of universal induction.

In this paper, we start from the concept of narrow machine learning methods as the result of specialization of universal induction [12], and show that practical meta-learning methods can be considered as the result of partial specialization of the universal induction. As the proof of concept, we develop a family of meta-inference methods in the form of deep neural networks and compare them on several tasks of different complexity. These methods differ in the completeness of specialization of the universal induction and range from learning discriminative models to learning task-specific genetic operators for genetic algorithms.

The main contribution of this paper is the framework, in which training discriminative models, learning to learn by gradient descent, and learning domain-specific crossover operators in genetic algorithms are represented as particular cases of specialization of universal induction. Deep learning models developed to demonstrate and verify these ideas can be considered as the minor contribution.

## 2   Background

The presented work is conceptually based on our two previous research directions. The first one is the theory of universal induction specialization [12]. The second one is implementation of the universal induction in the form of probabilistic programming languages with optimization queries (e.g. implemented in the form of simulated annealing and genetic programming) [13].

Solomonoff induction can be considered as the full Bayesian inference method, which utilizes a Turing-complete generative model that initially samples random program $z$ for universal Turing machine $U$ with universal priors $P_U(z)=2^{-l(z)}$ and then calculates its output $x=U[z]$ implying conditional probabilities $P_U(x|z)=1$ if $U[z]=x$ and 0 otherwise (one can also consider a stochastic universal Turing machine (UTM) defining smoother conditionals $P_U(x|z)$). In these settings, Bayesian inference can be performed, e.g. the marginal probability can be computed as

$$P_U(x) = \sum_z P_U(z)P_U(x|z) = \sum_{z:U[z]=x} 2^{-l(z)}. \qquad (1)$$

One can consider the generalized form of universal induction that takes as input an arbitrary machine $\mu$ that can be both universal and not universal. The machine accepts some $z$ treated as hidden variables. The task of induction is to calculate posterior distribution $P_\mu(z|x)$ or its maximum $z^*$. Machine $\mu$ can be treated as a generative model since it constructs (or samples in accordance with its likelihood function) $x$ using $z$: $x=\mu[z]$. We assume that some prior probability distribution over $z$ is also given making $\mu$ a probabilistic generative model.

Inference with the use of generative models consists in calculation of

$$P_\mu(z|x) = \frac{P_\mu(x|z)P_\mu(z)}{\sum_z P_\mu(x|z)P_\mu(z)} \text{ or} \qquad (2a)$$

$$z^* = \arg\max_z P_\mu(z|x) = \arg\max_z P_\mu(x|z)P_\mu(z). \qquad (2b)$$

E.g. if $\mu=U$ is UTM, then $z^* = \arg\max_z P_U(x|z)P_U(z) = \arg\max_{z:U[z]=x} 2^{-l(z)}$.

Usage of generative models encounters some difficulties since these models start from priors over models or hidden variables, and generate observables through nontrivial stochastic computations, so it is necessary to somehow guess appropriate values of hidden variables, model parameters, or even model structure that will produce real observations. That is, one should sum out $z$ or enumerate all values of $z$ in (2).

One can introduce the procedure of calculating (2) explicitly. Let us consider some search procedure $S(\mu, x)$ that takes machine/model $\mu$ as input, and returns the most probable $z^*$ or calculates $P_\mu(z|x)$. This procedure will correspond to a form of generalized universal induction.

While generative models allow for calculating any conditionals and marginals, but through intensive computations, discriminative or descriptive models directly and efficiently compute posterior probabilities or sample values of target or hidden variables. In the Bayesian approach, it is typical to construct a (variational) approximation to the posterior distribution specified by a generative model in the form of a discriminative model belonging to some family allowing efficient inference. That is, some machine $\nu$ is constructed such that $\nu[x] \approx z^*$ or $\nu[x] \approx P_\mu(z|x)$ depending on settings.

One can consider the problem of constructing a discriminative model given a generative model as the problem of specialization of the program $S$ performing universal induction w.r.t. its first parameter $\mu$. Indeed, the result of specialization of some program w.r.t. one of its parameters is the efficient version of this program with the fixed value of this parameter.

As the result of specialization of generalized universal induction procedure $S(\mu, x)$, one will get program $\nu = spec[S,\mu]$ such that
$$(\forall x)\nu[x] = S(\mu,x).$$

That is, *discriminative models are the results of complete specialization of the universal induction w.r.t. corresponding generative models*. One can also consider the problem of simultaneously learning machines $\mu$ and $\nu$ given some data that yields a sort of universal autoencoders [12].

Precise complete specialization is impossible in the case of a Turing-complete generative model. It is also doubtful that one can construct an approximate inversion $\nu \approx U^{-1}$, which will directly (without search) produce good enough programs given their outputs. Nevertheless, one can still hope to specialize $S$ w.r.t. $U$, i.e. to construct more efficient informed search method that takes $x$ as input and uses it to search for best $z$ taking the structure of $U$ and content of $x$ into account.

Recently we implemented $S$ as the simulated annealing and genetic programming search engine over probabilistic program traces [14]. Indeed, the idea to use genetic programming as the search method in universal induction is rather old [15] and well-known [16]. This leads us to the idea to specialize such a meta-heuristic method, i.e. to learn problem-specific and data driven genetic operators. It should be emphasized that learning such problem-specific genetic operators and constructing discriminative models have essentially the same meaning of specialization of universal induction, although the result of such specialization has rather different forms.

In this work, we don't do this within the probabilistic programming framework and just verify the very idea of learning genetic operations, but keep in mind that any fitness function can be defined as an optimization query in PPL. We represent a "genetic operator" (crossover and mutation) as a (deep) feedforward neural network that takes two candidate solutions and the values of parameters of the fitness function and learns to produce new candidate solution. Thus, more technically related works are the works on meta-learning in neural networks. For example, the classical work [10] is devoted to learning the learning strategy in the supervised learning settings. The more recent work [8] extends this result on the reinforcement learning settings. The work [7] is devoted to the problem of learning to learn by gradient descent by gradient descent. These works consider the problem of learning how to iteratively improve one candidate solution. One can think of our results loosely as the generalization of these methods to the arbitrary number (starting from zero) of candidate solutions.

The work [17] devoted to the "compilation" of probabilistic programs (generative models) into discriminative deep networks is also conceptually related. It should be pointed out that compilation is the particular form of specialization (namely, specialization of an interpreter w.r.t. a given program in accordance to Futamura-Turchin projections [20]). Thus, what is done by the authors is precisely a form of loose specialization of generalized universal induction w.r.t. a given program that we described earlier [12] (and which the authors seem not familiar with). Neural networks are used as a trainable proposal distribution, i.e. they again modify one given candidate solution.

One can also see a connection between our work and the idea of 'magician systems' described by Ben Goertzel in [18]: "Magician systems may thus be viewed as a kind of "generalized genetic algorithm," where the standard crossover operator is replaced by a flexible, individualized crossover operator… this is also precisely the type of dynamical system we need to use to model the brain/mind." Although the motivation and technical details of our work are completely different, we find this convergence of ideas quite interesting.

## 3   Models

Consider the task in which known family of fitness functions $f(\mathbf{x}|\boldsymbol{\theta})$ is given, and the goal is to find optimum $\mathbf{x}^*$ for given $\boldsymbol{\theta}$:

$$\mathbf{x}^*(\boldsymbol{\theta}) = \arg\min_{\mathbf{x}} f(\mathbf{x}|\boldsymbol{\theta}). \qquad (3)$$

Operation 'argmin' veils some computation that takes $\boldsymbol{\theta}$ as input and returns $\mathbf{x}^*$ as output. Such computations can vary from the completely uninformed random search to the direct calculation of $\mathbf{x}^*$ using explicit solution for a specific $f$.

We calculate (3) using different procedures:
- Blind search that randomly samples values of $\mathbf{x}$ and keeps track of the best value;
- Traditional genetic algorithms that perform uninformed meta-heuristic search in the space of $\mathbf{x}$ without taking $\boldsymbol{\theta}$ into account;
- Deep feedforward neural network that is trained to directly produce $\mathbf{x}^*(\boldsymbol{\theta})$ taking $\boldsymbol{\theta}$ as input: $\text{Net}_D(\boldsymbol{\theta}) \rightarrow \mathbf{x}^*$ working similar to discriminative models in pattern recognition;
- Genetic algorithms with trainable crossover operator represented in the form of deep feedforward network that takes two candidate solutions $\mathbf{x}_1$ and $\mathbf{x}_2$ and parameters $\boldsymbol{\theta}$ and produces new candidate solution $\mathbf{x}'$: $\text{Net}_{GA}(\mathbf{x}_1, \mathbf{x}_2, \boldsymbol{\theta}) \rightarrow \mathbf{x}'$.

Here, blind search (BS) and genetic algorithms (GA) are considered as general non-specialized search methods, while $\text{Net}_D$ and $\text{Net}_{GA}$ are considered as the result of different degree of specialization of general methods since they are trained to optimize a certain class of fitness-functions.

GAs were run on populations of small sizes (e.g. 10 survived species per population producing 20 children) to emphasize the role of recombinations. One step of blind search consisted in randomly sampling the same number of candidate solutions as the number of children in each population in GAs. The speed of mutations in GAs was adjusted to produce better results. In the case of $\text{Net}_{GA}$, mutations were applied to $\mathbf{x}_1$ and $\mathbf{x}_2$ before crossover instead of mutating the result of crossover as it is done in traditional GAs.

Both $\text{Net}_D$ and $\text{Net}_{GA}$ were fully connected feedforward networks with $H$ hidden layers with the number of neurons $100H$, $100(H-1)$, …, $100$ in the first, second, …, last layer correspondingly. Remarkably, networks with small number of layers produced (considerably) less precise solutions especially for the tasks of higher dimension. Here, we will show the results for *H*=5, because further increase of the network size leads to minor improvement of the precision.

Training of network parameters was performed by randomly sampling values of $\theta$ and using $|\text{Net}_D(\theta)-\mathbf{x}^*|^2$ and $|\text{Net}_{GA}(\mathbf{x}_1, \mathbf{x}_2, \theta)-\mathbf{x}^*|^2$ as components of the loss functions for the stochastic gradient descent. Values of $\mathbf{x}_1, \mathbf{x}_2$ for training $\text{Net}_{GA}$ were sample around $\mathbf{x}^*$, e.g. $|\mathbf{x}_{1,2} - \mathbf{x}^*|<1$.

## 4     Experiments

*Quadratic fitness functions*
Consider the following very simple task of optimization of quadratic functions. Let the fitness function be given in the form

$$f(\mathbf{x}|\mathbf{a},\mathbf{b}) = \mathbf{a}\mathbf{x}^2 + \mathbf{b}\mathbf{x}$$

where $\mathbf{x},\mathbf{a},\mathbf{b} \in R^N$ and $\mathbf{x}^2$ is element-wise, while multiplications are scalar products. Its minimum can be simply obtained analytically as $\mathbf{x}^*=-\mathbf{b}/(2\mathbf{a})$, where division is also element-wise. However, this task is not that trivial for neural networks trained by examples and not well suited to perform division.

We trained our models to produce the value of $\mathbf{x}^*$ taking $\mathbf{a}$, $\mathbf{b}$ or $\mathbf{a}$, $\mathbf{b}$ and $\mathbf{x}_1$, $\mathbf{x}_2$ as input, where $\mathbf{x}_1$, $\mathbf{x}_2$ are imprecise candidate solutions. That is, the network produces some $\mathbf{x}_i$ for randomly chosen task $(\mathbf{a}_i, \mathbf{b}_i)$ and the loss function $|\mathbf{x}_i-\mathbf{x}_i^*|^2$ is used to train the network using stochastic gradient descent. Random tasks were generated sampling $\mathbf{a}$~uniform(0.1, 1.1), $\mathbf{b}$~uniform(−1, 1).

Fig. 1 shows precision of solutions $|f(\mathbf{x}_{sol})-f(\mathbf{x}^*)|$ obtained by different methods depending on the number of iterations of search (i.e. generations in GAs) for *N*=5. The curve for $\text{Net}_D$ is constant since this method doesn't perform search. These curves are obtained by averaging over many (1000) optimization tasks.

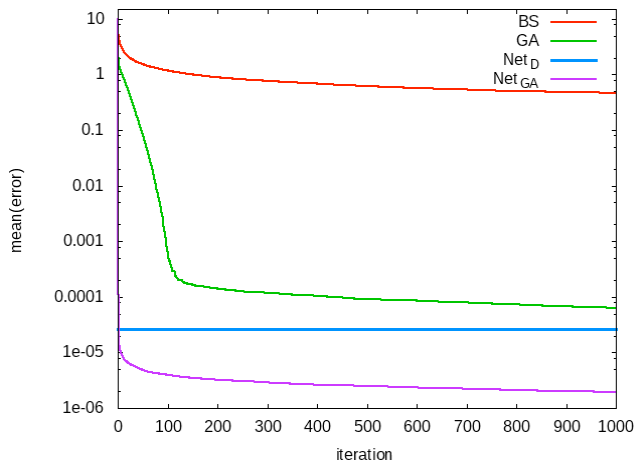

**Fig. 1.** Search efficiency in the task of quadratic function optimization

As it can be seen, blind search converges rather slowly, while other methods find reasonably good solutions quickly. This task appears to be simple for $Net_D$, although it cannot precisely represent division, so it produces imperfect solutions. $Net_{GA}$ has higher both convergence speed and precision in comparison to traditional GAs. Character of this curve seems to imply that $Net_{GA}$ relies more on the task parameters than on the candidate solutions and quickly produces candidate solutions of the same quality as $Net_D$. Nevertheless, its output is different for different input candidate solutions, and incremental improvement of the population of candidate solutions is achieved with its usage inside GA, so $Net_{GA}$ learns more complex mapping than $Net_D$.

*Linear equations*

Then, we compared the described models on the task of solving systems of linear equations:

$$\mathbf{Ax=b}$$

where $\mathbf{x}, \mathbf{b} \in R^N$, and $\mathbf{A}$ is $N \times N$ matrix.

Again, the models were trained on randomly generated tasks $\mathbf{A}_i$, $\mathbf{b}_i$ to produce $\mathbf{x}_i$ minimizing $|\mathbf{A}_i\mathbf{x}_i - \mathbf{b}_i|^2$. Random sampling was performed as $\mathbf{A}, \mathbf{b} \sim$ uniform$(-1,1)$, but rejecting tasks with solutions $\mathbf{x}^*$ such that $|\mathbf{x}^*|>6$.

This problem appeared to be considerably more difficult for neural networks. Fig. 2 shows the obtained averaged solution error $\|\mathbf{Ax}_{sol} - \mathbf{b}\|$ for $N=5$.

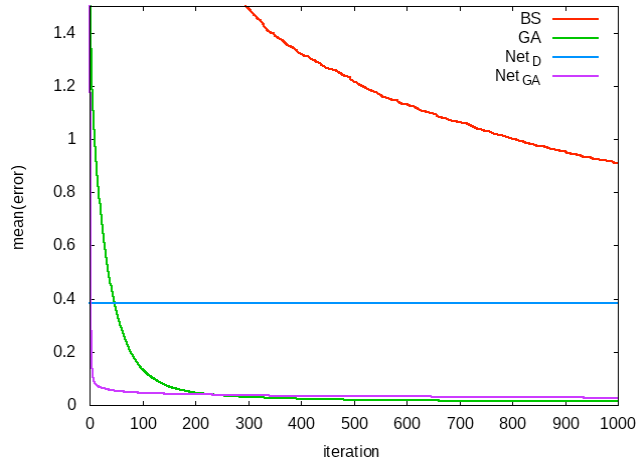

**Fig. 2.** Search efficiency in the task of solving linear equations

As it can be seen, $Net_D$ fails to learn good mapping from the space of parameters of linear equation systems to the space of their solutions, although it produces better results than achieved by blind search in a reasonable number of steps. Although this task is not NP-complete, and can be solved without search, it cannot be solved by linear algorithms, so this result is not surprising.

At the same time, $Net_{GA}$ solves this task in few iterations, i.e. much faster than traditional GAs. On the other hand, $Net_{GA}$ converges to slightly less precise solutions in average. Again, the reason might be that $Net_{GA}$ relies more on the task parameters than on the parent candidate solutions. That is, it learns the mapping, which is closer to complete specialization than to traditional crossover. This result is also reasonable taking into account that $Net_{GA}$ is trained to produce candidate solutions as close to the optimal solution as possible. Quite opposite, it might be surprising that the network learned to use the parent candidate solutions in addition to the task parameters. It should be mentioned that this effect is achieved easier when $Net_{GA}$ is trained on parents that are close to $\mathbf{x}^*$ (i.e. parents are not arbitrary, but contain some information about $\mathbf{x}^*$).

*Basic meta learning*

The last task we considered was the task of producing parameters of the optimal logistic regression model given the training set. That is, the weights and biases in the logistic regression network act as $\mathbf{x}$, while the training set for this model is considered as $\mathbf{\theta}$. Thus, the task was to learn the learning algorithm that maps training sets to the logistic regression parameters.

The patterns for the training sets were sampled from two Gaussians corresponding to two classes. Parameters of these Gaussians were generated randomly, but in such a way that the distance between centers of classes was between $2\sigma$ and $3\sigma$.

This task appeared to be very simple, and complete specialization $Net_D$ produced almost optimal solutions. For example, for the dimension of patterns $N=2$ and the size of training set $N_{train}=20$ (with random number of patterns per class) the averaged results for different number of hidden layers are shown in Table 1, while recognition rate of the logistic regression from sklearn library was 0.9837.

Table 1. Results of recognition by logistic regression produced by $Net_D$

|  | $H=1$ | $H=3$ | $H=5$ | $H=10$ |
|---|---|---|---|---|
| Recognition | 0.9335 | 0.9840 | 0.9845 | 0.9842 |

The same results were obtained for larger values of $N$. Unfortunately this straightforward approach to meta learning doesn't scale to the real pattern recognition problems. The main limitation consists in the usage of the whole training set as the input to $Net_D$ or $Net_{GA}$. More practical approach would be to pass patterns from the training set one by one to these networks, but then the networks should either be recurrent in order to be able to accumulate information from the patterns, or be trained in a very specific way to perform a sort of stochastic gradient descent step. Development of such models is beyond the scope of this paper, and is the topic of further work.

## 5 Discussion

Although our experiments were conducted on example of rather simple synthetic problems, they demonstrate the following ideas:
- There can be different degrees of specialization of general search procedures including complete and partial specialization, and the optimal degree of spe-cialization depends on the family of problems to be solved. In particular, there is a large set of models between generative and discriminative models, and approximating inference in generative models with discriminative models is not the only and sometimes not the best choice.
- One example of partial specialization is genetic algorithms with the trainable crossover operator that accepts not only two parent candidate solutions, but also the parameters of the task to be solved. Such specialized GAs can converge much faster than traditional GAs, and their performance can be much better than that of complete specialization.
- Such trainable crossover operators can be productively implemented in the form of deep neural networks at least for some families of tasks.

Although these conclusions are true in general, their significance for the real-world problems and AGI systems is still to be studied in detail. In particular, we conducted some additional experiments showing some limitations of the implemented form of trainable GAs.

First of all, it appeared that both $Net_D$ and $Net_{GA}$ work bad on the tasks outside the region of the training set, i.e. neural networks don't generalize well (at least in a traditional sense). For example, in the task of quadratic functions minimization, they don't learn the division operation enabling calculation of $x^*=-b/(2a)$ for any $a$ and $b$, but memorize this mapping for specific $a_i$, $b_i$ from the training set and interpolate it. This conclusion is consistent with some recent studies (e.g. [19]).

Then, we compared $Net_{GA}$ with the network that takes not two, but only one parent as input, and is also used inside the search procedure. Briefly speaking, we didn't observe considerable difference in their performance. Thus, our specialized search didn't benefit much from recombining two candidate solutions. We believe that it can benefit considerably (because GAs can be considerably better than simulated annealing or gradient descent in some tasks), but more complex tasks should be considered and/or less simplistic loss function should be used.

Indeed, we trained $Net_{GA}$ outside the search cycle. It was required to produce as good candidate solution as possible after single application to the random parents, while its usage within GA supposes its iterative application to the evolving population of solutions with the aim to find the optimal solution not immediately, but after a number of generations. Efficient approach to representing and optimizing such loss function is to be developed. One possibility is to represent the whole search process as a recurrent neural network to optimize it end-to-end.

Further development of this approach also consists in its application to arbitrary optimization queries in probabilistic programming. One lesson that we can learn from our study is that probabilistic programs cannot be "compiled" into feedforward neural networks in general case.

## Acknowledgements

This work was supported by Government of Russian Federation, Grant 074-U01.